\documentclass{article}

\usepackage{arxiv}

\usepackage[utf8]{inputenc} 
\usepackage[T1]{fontenc}    
\usepackage{hyperref}       
\usepackage{url}            
\usepackage{booktabs}       
\usepackage{amsfonts}       
\usepackage{nicefrac}       
\usepackage{microtype}      
\usepackage{graphicx}
\usepackage{amsmath}
\title{TGCN:Time Domain Graph Convolutional Network for Multiple Objects Tracking}

\author{
  Jie Zhang\\

}

\begin{document}
\maketitle

\begin{abstract}
Multiple object tracking is to give each object an id in the video. The difficulty is how to match the predicted objects and detected objects in same frames. Matching features include appearance features, location features, etc. These features of the predicted object are basically based on some previous frames. However, few papers describe the relationship in the time domain between the previous frame features and the current frame features.In this paper, we proposed a time domain graph convolutional network for multiple objects tracking.The model is mainly divided into two parts, we first use convolutional neural network (CNN) to extract pedestrian appearance feature, which is a normal operation processing image in deep learning, then we use GCN to model some past frames' appearance feature to get the prediction appearance feature of the current frame. Due to this extension, we can get the pose features of the current frame according to the relationship between some frames in the past. Experimental evaluation shows that our extensions improve the MOTA by 1.3 on the MOT16, achieving overall competitive performance at high frame rates.
\end{abstract}

\keywords{Computer vision \and Multiple object tracking \and Graph convolutional network}

\section{Introduction}
Multiple object tracking(MOT) is an active research problem in computer vision area. Due to recent progress in object detection\cite{7485869,Redmon_2016_CVPR}, tracking-by-detection\cite{7533003}has become the leading paradigm in MOT. Although recent years have witnessed rapid advancements in MOT, it is still a challenging task due to large changes of object motion features caused by pose, illumination, deformation and occlusion, etc.

One main issue for MOT problem is how to match the predicted objects and detected objects in same frames. These two kind of objects need features to represent them. But how to generate strong discriminative feature representation for predicted objects has attracted the attention of many scholars. For example, bounding box position and size are used for both motion estimation and data association. This method can not solve the problem of identity switches when two objects meet. Furthermore, appearance features can be added to improve accuracy.   

\begin{figure}[htp]
    \centering
    \includegraphics[width=10cm]{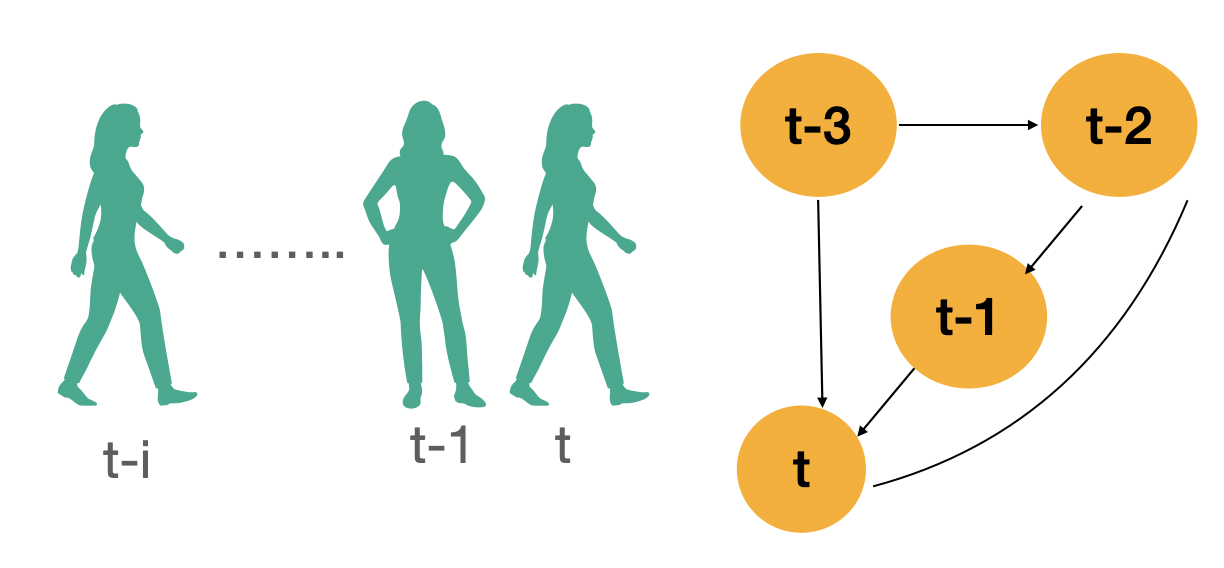}
    \caption{The relationship between frames}
    \label{fig:fig1}
\end{figure}

Since the actions such as walking are continuous and periodic in time domain, we try to use GCN to model the pose features in time dimension.See figure \ref{fig:fig1}. First of all, in order to make the t-1 frame have an impact on the t frame, we can add the connection relationship between the t-1 frame and the t frame as a priori knowledge. Furthermore, we add a data-driven probability model to characterize the relationship between t-i frames and t frames. After several layers of GCN, we can get the features of the current frame according to the relationship between some frames in the past.

\section{Related works}
\subsection{Tracking-by-detection}
Recently, many studies have employed detection for MOT to improve MOT results.Firstly, an independent detector is applied to all image frames to obtain likely pedestrian detections. Secondly, a tracker is run on the set of detections to perform data association, link the detections.Li et al.\cite{4479480} used a probabilistic model combining conventional tracking and object detection to track objects in low frame rate (LFR) video;  Grabner and Bischof \cite{1640768} used a real-time Adaboost feature selection model for object detection. This work reduced the computational complexity of Adaboost significantly, but because of the limitations of the number of weak classifiers, the accuracy of the detector was low.Online detectors are more adaptable, and are able to track a specific target amongst many objects from the same class. However, these classifiers perform worse than offline detectors in terms of accuracy, and training data extracted.

Due to recent progress in object detection\cite{7485869,Redmon_2016_CVPR}, tracking-by-detection has become the leading paradigm in MOT. A. Bewley et al.\cite{7533003}proposed that bounding box position and size are used for both motion estimation and data association. This method can not solve the problem of identity switches when two objects meet. In order to reduce the identity switches situation, Nicolai Wojke et al. \cite{8296962}integrate appearance information to improve the performance of SORT. Due to this extension they are able to track objects through longer periods of occlusions, effectively reducing the number of identity switches. However, it is well known that the motion features such as pose feature at t time are not necessarily the closest to t-1, but may be related to t-i, where i is less than t. 

\subsection{Graph Convolutional Networks}
As an extension of CNNs from regular grid to irregular graph, graph convolutional networks (GCNs) \cite{kipf2017semisupervised,2018graph,hu2019hierarchical,8953909,li2018adaptive}have been demonstrated very effectively in graph representation and learning.Kipf et al. \cite{kipf2017semisupervised}propose to develop a simple Graph Convolutional Network (GCN) for graph semi-supervised learning. Hamilton et al. \cite{hamilton2018inductive} present
a general inductive representation and learning framework for the representations of unseen nodes.Velickovi et al. \cite{2018graph}propose Graph Attention Networks (GATs) for graph based semi-supervised learning.The core GCNs is to conduct graph node representation and labeling by propagating messages on graph structure.Recently, knowledge graph convolutional networks have been developed for multi-label recognition.Chen et al. \cite{chen2019multilabel}propose multi-label GCN (ML-GCN) for image multi-label recognition. ML-GCN aims to learn interdependent object classifiers by employing GCN learning on an object label graph which encodes the correlation information among different labels. 

Inspired by these works, we propose to construct
an time domain graph to model the relationship
information among different frames and then employ a noval time domain graph convolutional network (TGCN) for multiple objects tracking.

\section{Proposed model}
\begin{figure}[htp]
    \centering
    \includegraphics[width=15cm]{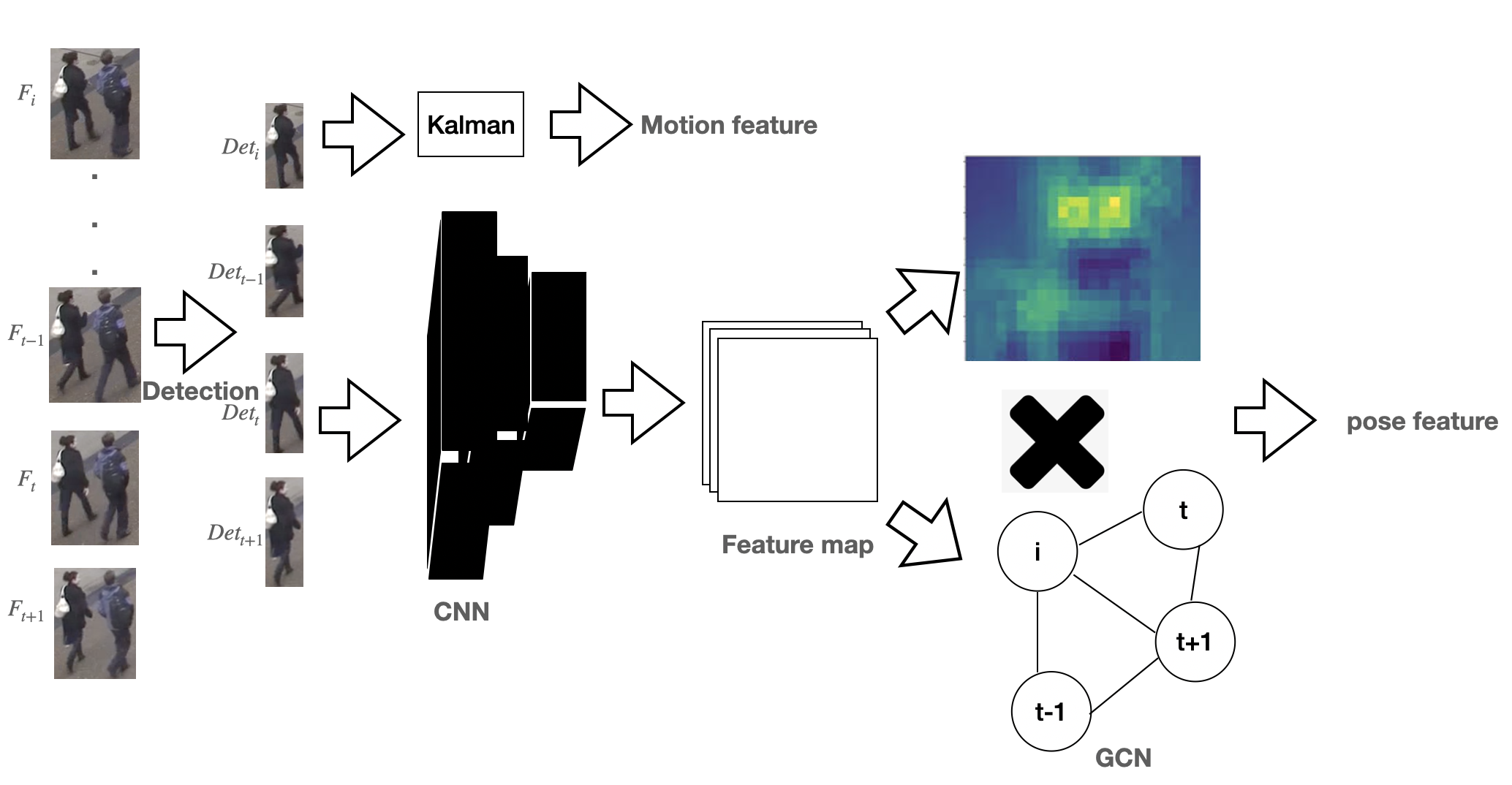}
    \caption{The relationship between frames}
    \label{fig:fig2}
\end{figure}

In this section, we present our time domain graph convolutional network (TGCN) for multiple objects tracking tasks. The overall framework of TGCN is shown in figure \ref{fig:fig2} which contains four main parts,1) CNN based object detection method, 2)Motion state estimation,
3) TGCN based tracking pose feature extraction, 4) association between the predicted Kalman states and pose distance. In the following, we present the detail of these modules.

\subsection{CNN based object detection method}
To capitalise on the rapid advancement of CNN based detection, we utilise the YOLOv5 detection framework.YOLOv5 is an end-to-end framework that only consists of one stages.And it is the best real-time algorithm in the field of object detection.Using the GPU of V100 yolov5 infers within 20 ms on a single image (batch size 1), that makes real-time deployment possible. Additionally, the AP of yolov5 can reach 0.5 on coco dataset.In this paper, we compare the detection performance of faster RCNN and yolov5 as detectors, and the tracking performance based on them.See Table\ref{tab:table1}. As we are only interested in pedestrians we ignore all other classes and only pass person detection results with output probabilities greater than 0.5 to the tracking framework.

\begin{table}[htp]
 \caption{Comparison of tracking performance by switching
the detector component}
  \centering
  \begin{tabular}{llll}
    \toprule
    Tracker  &   detector       & detection   & tracking \\
    \midrule
    deepSort &  faster RCNN(X101-FPN)  & 43  & 61.4    \\
    deepSort &   YOLOv5x         & 49.2   &62.3      \\
    Ours     &   faster RCNN(X101-FPN) &  43  & 62.1     \\
    Ours     &   YOLOv5x         & 49.2      & 62.7      \\
    \bottomrule
  \end{tabular}
  \label{tab:table1}
\end{table}

In our experiments, we verified that the detection quality has a significant impact on tracking performance when comparing the yolov5x detections to faster RCNN(X101-FPN) detections. 

\subsection{Motion state estimation}
In motion estimation part, we add velocity vector based on deep sort.Therefore, our Kalman filter attempts to predict six parameters, such as $(x,y,w,h,vx,vy)$, that contains the bounding box coordinates$(x,y,w,h)$and velocity vector $(vx,vy)$.We increase the velocity vector mainly to reduce the mismatching based on appearance features in deepsort. For example, when two people in similar colors meet face to face, the tracker is likely to fail.

Firstly, we initialize our Kalman filter with the detection results of the start frame. Secondly, we use the Kalman filter to predict the position and velocity of the next frame. Thirdly, we use the detection results of the next frame to correct the Kalman filter.Repeat the second steps and thirdly steps.In this paper, We compare the Identity switches of the algorithm with velocity vector and non velocity vector.See Table\ref{tab:table2}.

\begin{table}[htp]
 \caption{Comparison of tracking performance by switching
velocity vector}
  \centering
  \begin{tabular}{lll}
    \toprule
    Tracker  &   add velocity         & ID switch \\
    \midrule
    deepSort &  no           & 781     \\
    deepSort &   yes         & 758        \\
    Ours     &   no          & 776       \\
    Ours     &   yes         & 751           \\
    \bottomrule
  \end{tabular}
  \label{tab:table2}
\end{table}

\subsection{TGCN based tracking pose feature extraction}
There are two problems to be solved. First, we need to use CNN to express the appearance feature of pedestrians in a single frame. Second, we need to use GCN to describe the spatial-temporal relationship between the past $i$ frames. On the basis of the appearance feature plus the spatial-temporal features, we can get the pose feature. 

Convolutional neural networks have achieved excellent performance in image processing these years. We can use any CNN base models to learn the features of each pedestrian image. In our experiments, our appearance feature extraction model use ResNet-101 to extract pedestrian image features. This is mainly because the residual network will not cause gradient disappearance with the deepening of the network. We randomly resize each training image into 448 × 448 resolution and then we can obtain 2048 × 14 × 14 feature maps from the $conv5_x$ layer. Next, we use the global max pooling to process the feature map, and we can get the new feature map M, And 2048 × 1 is its dimension. we define the time window is $C$. Therefore, the input to GCN network is $X^{0}$, and $C$ × 2048 is its dimension. We adopt a L-layer GCN module for spatial-temporal relationship representation and learning. Formally, given $A\in{R^{C*C}}$,and the l-layer representation $X^{(l)}\in{R^{C*d}}$, we propose to conduct
the layer-wise propagation as
\begin{equation}
X^{(l+1)} = \sigma{(AX^{(l)}W)}
\end{equation}
where $l=0,1,...,L-1$.And A indicates the correlation matrix. $\sigma(.)$ is an activation function, such as ReLU function. W denotes the layer-wise trainable
weight matrix. After GCN, we can get the output $X^{(L)}, and $1 × $C$ is its dimension.

Graph convolutional network works by propagating information between nodes based on the correlation matrix.
Thus, it is crucial to construct a correlation matrix. Normally, we use adjacent matrix to represent the topology structure of a graph. First of all, according to prior knowledge, we think that t-1 frame is connected to t frame, which is a unidirectional connection. Therefore, we first define the matrix Q:
\begin{equation}
Q = 
\begin{bmatrix}
0 & 1  & 0 & \cdots   & 0   \\
0 & 0  & 1 & \cdots   & 0  \\
\vdots & \vdots & \vdots & \ddots   & \vdots  \\
0 & 0  & 0 & \cdots\  & 0  \\
\end{bmatrix}
\end{equation}
And then we define the matrix P:
\begin{equation}
P = 
\begin{bmatrix}
p_{0,0} & p_{0,1}  & p_{0,2} & \cdots   & p_{0,t}   \\
p_{1,0} & p_{1,1}  & p_{1,2} & \cdots   & p_{1,t}  \\
\vdots & \vdots & \vdots & \ddots   & \vdots  \\
p_{t,0} & p_{t,1}  & p_{t,2} & \cdots\  & p_{t,t}  \\
\end{bmatrix}
\end{equation}

$p_{i,j}$ represents the probability that $j$ frame feature is closest to $i$ frame feature.We consider the adjacency matrix A:
\begin{equation}
A = Q+P
\end{equation}
Finally, we define the loss function L:
\begin{equation}
L = (M_{t+1} - {X_{t}}^{(0)}* X_{t}^{(L)} )^{2}
\end{equation}
\subsection{Association}
A conventional way to solve the association between the predicted Kalman states and pose feature is to build an assignment problem that can be solved using the
Hungarian algorithm. Into this problem formulation we integrate motion and pose information through combination of two appropriate metrics.
To incorporate motion information $(x,y,w,h)$, the (squared) Mahalanobis distance $d_1$ is mostly identical to the original formulation in deepsort. To further measure the distance of the velocity vector, we define $d_2$:
\begin{equation}
d_2=1-cos(\theta)= 1-\frac{{v_1}{v_2}}{\left\|v_1\right\|\left\|v_2\right\|}
\end{equation}
In order to measure the similarity of pose feature, we
integrate $d_3$ into the assignment problem. And it is mostly identical to the original formulation in deepsort.In order to combine the above distances, we define the $D$ of Hungarian matching as
\begin{equation}
D=\lambda_1(d_1)+\lambda_2(d_2)+(1-\lambda_1-\lambda_2)(d_3)
\end{equation}

\section{Experiments}
We evaluate the performance of our tracking implementation on a diverse set of testing sequences as set by the MOT16
benchmark database which contains both moving and
static camera sequences. Evaluation is carried out according to the following metrics: MOTA(Multi-object tracking accuracy), MOTP(Multi-object tracking precision), MT(number of mostly tracked trajectories), ML(number of mostly lost trajectorie),ID swith(number of times an ID switches to a different previously tracked object), FP(number of false detections), FN( number of missed detections),FM(Number of times a track is interrupted
by a missing detection). Table\ref{tab:table3} compares the proposed method TGCN with several other baseline trackers. Compared to these other methods, TGCN achieves competitive score. Thanks to stronger detectors, our method  successfully reduce the FP/FN/FM numbers.  At the same time, the  velocity vector and pose feature help reduce the number of ID switches compared with deep sort method.

\begin{table}[htp]
 \caption{Comparison of proposed method TGCN with several other baseline trackers}
  \centering
  \begin{tabular}{lllllllll}
    \toprule
    Method  & MOTA & MOTP & MT & ML & ID switch & FM & FP & FN \\
    \midrule
    EAMTT  & 52.5 & 78.8 & 0.190 & 0.349 & 910 & 1321 & 4407 & 81223 \\
    POI  & 66.1 & 79.5 & 0.34 & 0.208 & 805 & 3093 & 5061 & 55914 \\
    Deepsort  & 61.4 & 79.1 & 0.328 & 0.182 & 781 & 2008 & 12852 & 56668 \\
    TGCN  & 62.7 & 79.3 & 0.316 & 0.205 & 751 & 1976 & 7101 & 55581 \\
    \bottomrule
  \end{tabular}
  \label{tab:table3}
\end{table}

\section{CONCLUSION}
This paper proposes a novel time domain graph convolutional network model for multiple objects tracking. TGCN employs a novel data-driven Graph convolutional architecture for time domain pose representation and learning. Experimental results on benchmarks MOT16 demonstrate that TGCN is a strong competitor to state-of-the-art multiple objects tracking approaches. In the future, we will adopt TGCN to address some other computer vision tasks, such as Re-ID.

\bibliographystyle{unsrt}  


\begin{thebibliography}{1}

\bibitem{7485869}
S. Ren and K. He and R. Girshick and J. Sun.
\newblock Faster R-CNN: Towards Real-Time Object Detection with Region Proposal Networks.
\newblock In IEEE Transactions on Pattern Analysis and Machine Intelligence, vol. 39, no. 6, pp. 1137-1149, 1 June 2017, doi: 10.1109/TPAMI.2016.2577031.

\bibitem{Redmon_2016_CVPR}
Joseph Redmon, Santosh Divvala, Ross Girshick, Ali Farhadi.
\newblock You Only Look Once: Unified, Real-Time Object Detection.
\newblock In IEEE Conference on Computer Vision and Pattern Recognition (CVPR), 2016, pp. 779-788.

\bibitem{7533003}
A. Bewley, Z. Ge, L. Ott, F. Ramos and B. Upcroft.
\newblock Simple online and realtime tracking.
\newblock In 2016 IEEE International Conference on Image Processing (ICIP), Phoenix, AZ, 2016, pp. 3464-3468, doi: 10.1109/ICIP.2016.7533003.

\bibitem{4479480}
Y. Li, H. Ai, T. Yamashita, S. Lao and M. Kawade.
\newblock Tracking in Low Frame Rate Video: A Cascade Particle Filter with Discriminative Observers of Different Life Spans.
\newblock In Tracking in Low Frame Rate Video: A Cascade Particle Filter with Discriminative Observers of Different Life Spans.

\bibitem{1640768}
H. Grabner and H. Bischof.
\newblock On-line Boosting and Vision.
\newblock In 2006 IEEE Computer Society Conference on Computer Vision and Pattern Recognition (CVPR'06), New York, NY, USA, 2006, pp. 260-267, doi: 10.1109/CVPR.2006.215.

\bibitem{8296962}
N. Wojke, A. Bewley and D. Paulus.
\newblock Simple online and realtime tracking with a deep association metric.
\newblock In 2017 IEEE International Conference on Image Processing (ICIP), Beijing, 2017, pp. 3645-3649, doi: 10.1109/ICIP.2017.8296962.

\bibitem{kipf2017semisupervised}
Thomas N. Kipf and Max Welling.
\newblock Semi-Supervised Classification with Graph Convolutional Networks.
\newblock In arXiv preprint arXiv:1609.02907, 2017.

\bibitem{2018graph}
Petar Veličković and Guillem Cucurull and Arantxa Casanova and Adriana Romero and Pietro Liò and Yoshua Bengio.
\newblock Graph Attention Networks.
\newblock In arXiv preprint arXiv:1710.10903, 2018.

\bibitem{hu2019hierarchical}
Fenyu Hu and Yanqiao Zhu and Shu Wu and Liang Wang and Tieniu Tan.
\newblock Hierarchical Graph Convolutional Networks for Semi-supervised Node Classification.
\newblock In arXiv preprint arXiv:1902.06667, 2019.

\bibitem{8953909}
B. Jiang, Z. Zhang, D. Lin, J. Tang and B. Luo.
\newblock Semi-Supervised Learning With Graph Learning-Convolutional Networks.
\newblock In 2019 IEEE/CVF Conference on Computer Vision and Pattern Recognition (CVPR), Long Beach, CA, USA, 2019, pp. 11305-11312, doi: 10.1109/CVPR.2019.01157.

\bibitem{li2018adaptive}
Ruoyu Li and Sheng Wang and Feiyun Zhu and Junzhou Huang.
\newblock Adaptive Graph Convolutional Neural Networks.
\newblock In arXiv preprint arXiv:1801.03226, 2018.

\bibitem{hamilton2018inductive}
William L. Hamilton and Rex Ying and Jure Leskovec.
\newblock Inductive Representation Learning on Large Graphs.
\newblock In arXiv preprint arXiv:1706.02216, 2018.

\bibitem{chen2019multilabel}
Zhao-Min Chen and Xiu-Shen Wei and Peng Wang and Yanwen Guo.
\newblock Multi-Label Image Recognition with Graph Convolutional Networks.
\newblock In arXiv preprint arXiv:1904.03582, 2019.


\end{thebibliography}

\end{document}